\documentclass[runningheads]{llncs}

\usepackage[utf8]{inputenc}

\usepackage{hyperref}
\usepackage{multicol}
\usepackage{multirow}
\usepackage{xcolor}
\usepackage{graphics}

\begin{document}
\title{GERNERMED++: Robust German Medical NLP through Transfer Learning, Translation and Word Alignment}

\author{Johann Frei\inst{1}\orcidID{0000-0003-0323-0904} \and
Ludwig Frei-Stuber\inst{2} \and
Frank Kramer\inst{1}\orcidID{0000-0002-2857-7122}
}

\authorrunning{Frei et al.}

\institute{
IT-Infrastructure for Translational Medical Research, University of Augsburg, 86159 Augsburg, Germany\\
\email{firstname.lastname@informatik.uni-augsburg.de}
\and
Institute and Outpatient Clinic for Occupational, Social and Environmental Medicine, 80336 Munich, Germany\\
\email{firstname.lastname@med.uni-muenchen.de}
}

\maketitle

\begin{abstract}
We present a statistical model for German medical natural language processing trained for named entity recognition (NER) as an open, publicly available model. The work serves as a refined successor to our first GERNERMED model which is substantially outperformed by our work. We demonstrate the effectiveness of combining multiple techniques in order to achieve strong results in entity recognition performance by the means of transfer-learning on pretrained deep language models (LM), word-alignment and neural machine translation. Due to the sparse situation on open, public medical entity recognition models for German texts, this work offers benefits to the German research community on medical NLP as a baseline model.
Since our model is based on public English data, its weights are provided without legal restrictions on usage and distribution.
The sample code and the statistical model is available at:\\\url{https://github.com/frankkramer-lab/GERNERMED-pp}

\end{abstract}

\section{Introduction}

Extraction and processing of key information from medical notes and doctors' letters poses a common challenge in advanced digitization of health care systems. In particular, research-oriented data mining of non-research-centric data sources (often referred to as \textit{second use}) often requires expensive data harmonization processes in order to transform unstructured or semi-structured data into strictly structured, uniform data representations such as HL7 or FHIR. While manually solving these processes can be carried out for document analysis on certain studies, it is rendered impractical for large scale text analysis on legacy data or processing day-to-day clinical data.

Handling heterogeneous data from text-based documents is a central subject of natural language processing. In recent years deep learning-inspired approaches have been applied successfully to tackle various NLP tasks effectively. However, training deep language models requires proper datasets in regards to aspects like corpus size, annotation work, data diversity and overall dataset quality, in order to retrieve well-performing models. In medical NLP, obtaining such annotated datasets remains rather difficult because the use and publication of medical data is highly restricted for the reasons of privacy and country-dependent data protection legislation. Even though medical datasets have been published in English, such datasets for German texts in contrast are still frequently unavailable to external researchers.

In this paper, we propose an approach of combining multiple ideas to obtain a German medical NLP model:
\begin{itemize}
    \item \textbf{Translation}: The state of German medical corpora is limited and the use of internal datasets for training and publication of models is legally disputed in regards to the fact that the possibility of privacy-concerned training data extraction cannot be excluded. In contrast, medical datasets in English have already been published and therefore, neural machine translation (NMT) can be applied to obtain German data from English datasets. 
    \item \textbf{Annotation Projection}: Annotation of large corpora is crucial for supervised learning and determines the quality of the final performance of the model. However the cost of obtaining gold-standard annotations from scratch opposes the practical feasibility. Given our set of NMT-based German data, word alignment estimation can be used to project token-level annotations from English data to German data without manual intervention.
    \item \textbf{Transfer-Learning through Model Fine-Tuning}: To further improve the downstream performance of the NLP model under the constraints of our small, task-specific dataset, a larger, pre-trained German LM is used for advanced semantic, context-aware feature extraction and further fine-tuning.
\end{itemize}

\subsection{Related Work}
In the recent decade, in particular in the last five years, the field of natural language processing has been radically transformed by the use of data-driven, neural methods that are able to surpass previous state-of-the-art performances. The introduction of the attention-based transformer model\cite{vaswani_attention_2017} in the field of NLP lead to various follow-up works such as BERT\cite{devlin_bert_2018} and similar deep language models that are trained and applied on domain-specific contexts\cite{peng_transfer_2019,rasmy_med-bert_2021,lee_biobert_2019,alsentzer_publicly_2019,beltagy_scibert_2019,li_fine-tuning_2019}. All these domain-specific works share in common that their research focus lies primarily on English application and use.

The training of novel transformer-based German NLP models requires proper datasets with respect to size and quality. In purely supervised scenarios, this also includes the need for gold-standard annotation labels. While several works with internal datasets exist, their datasets are not shared among the research community and remain undisclosed\cite{wermter_annotated_2004,fette_information_2012,bretschneider_identifying_2013,toepfer_fine-grained_2015,lohr_operative_1992,kreuzthaler_detection_2015,roller_fine-grained_2016,cotik_negation_2016,krebs_semi-automatic_2017,hahn_3000pa-towards_2018,minarro-gimenez_quantitative_2019,konig_knowledge-based_2019,schafer_towards_2020}, and thus this presents major hurdles for open research and independent reproducibility. The situation on public, English datasets is more convenient and several large datasets like MIMIC-III\cite{pollard_mimic-iii_2016} or the i2b2 challenges with datasets such as the n2c2 2018 dataset\cite{henry_2018_2020} have been published, as well as the multilingual Mantra GSC\cite{kors_multilingual_2015} dataset from the biomedical domain. Only in recent few years has the German medical NLP research community addressed this issue and developed novel German medical datasets that are publicly accessible as foundation for future NLP work.\cite{borchert_ggponc_2020,kittner_annotation_2021}. Regarding the GGPONC\cite{borchert_ggponc_2020}, an updated iteration has been presented\cite{borchert_ggponc_2022}.

With regards to novel German medical NLP systems, commercial software like \textit{Averbis Health Discovery}\cite{noauthor_averbis_nodate}\footnote{\url{https://averbis.com/de/health-discovery/}} and \textit{German Spark NLP for Healthcare}\cite{noauthor_detect_nodate}\footnote{\url{https://nlp.johnsnowlabs.com/2021/03/31/ner_healthcare_de.html}} are proprietary and require licenses. As an exception, \textit{mEx}\cite{roller_mex_2018} is freely available, but the model weights must be requested and legal restrictions on the models' usage are imposed. For German medical NER tasks, only few public, open neural models are available to the best of our knowledge, such as \textit{GGPONC}\cite{borchert_ggponc_2022} and \textit{GERNERMED}\cite{frei_gernermed_2022}.

\section{Methods}

\subsection{Dataset Acquisition}
The dataset retrieval pipeline for German texts follows the approach proposed in GERNERMED\cite{frei_gernermed_2022}: As a starting point, the \textit{2018 n2c2 shared task on ADE and medication extraction in EHR} dataset serves as an English source dataset of medical entities from anonymized electronic health records. The source dataset is decomposed into sentences as the initial preprocessing step. During that process, text spans that have been replaced with an anonymized identifier text bracket by the editors of the source dataset are detected and replaced with randomized synthetic data from the \textit{Faker} Python module in order to reduce the number of irregular text occurrences while updating the initial annotation span indices accordingly. For instance, this includes text entities like first and family name, dates and postal addresses.

We apply the publicly available FAIRseq \textit{transformer.wmt19.en-de}\cite{ott_fairseq_2019} NMT model for sentence-wise automatic translation, which features a transformer-based neural model for translating sentences from English to German. Since the annotation information from the source dataset cannot be directly preserved for German sentences, the reconstruction of the annotation spans for the translated sentences can be estimated by the means of a bitext word alignment as a postprocessing step. In contrast to the approach in GERNERMED, we refine the word alignment estimation step in regard to the following aspects:
\begin{itemize}
    \item \textbf{Improved Tokenization}: The tokenization of sentences for the word alignment differs from modern tokenizers that generate sub-word-level tokens optimized through techniques such as byte pair encoding schemes. Most word alignment methods operate on word-level tokenization with whitespace-based token splitting. In order to reduce the number of misaligned words, we further refined the word-level tokenization by separating punctuation from words instead of only relying on tokenization splits on whitespace characters.
    \item \textbf{Word Alignment Technique}: In NLP bitext word alignment is the task of determining the semantic correspondence between words from a bilingual sentence pair consisting of the source and translated sentence. In previous work, the \textit{Fast\_Align}\cite{dyer_simple_2013} implementation has been used for establishing such correspondences. It uses the IBM 2 alignment model for alignment estimation in a purely unsupervised fashion. While there are also other models inspired from statistical machine translation\cite{och_systematic_2003,ostling_efficient_2016}, recent work has been done towards neural approaches\cite{jalili_sabet_simalign_2020,dou_word_2021}. For this work, we use the pre-trained model from \textit{Awesome-Align}\cite{dou_word_2021}. In short, the model tackles the task by encoding both sentences through a pretrained crosslingual language model in order to obtain contextualized word vector embeddings. Although the words of the sentence pairs largely differ with respect to their syntactic and linguistic features, the implementation makes use of the assumption that corresponding words are similar in terms of their word vectors in embedding space in order to find the word correlations in each sentence.
\end{itemize}

After the translation of the sentences, applying the word alignment estimation on the set of sentence pairs given the refinements for tokenizer and word alignment yields essential information on the relationship between the annotation spans of the English entity labels and their German counterparts. As a minor disadvantage of the common \textit{Pharaoh} alignment format, the difference in annotation granularity cannot be preserved completely on character level. Even though the annotation spans of the source dataset are provided as character-level indices, the word-level tokenization restricts the ability to reconstruct sub-word-level annotation spans in the German target data when the backprojection of the word-level indices from the word alignment estimation onto the character-level indices of the target sentence text string is evaluated.

\subsection{Entity Recognition Training}

The training of our entity recognition model employs the entity recognition parser from the \textit{SpaCy} library which follows a transducer-based parsing approach with a BILOU scheme (\textit{\textbf{B}egin},\textit{\textbf{I}nside},\textit{\textbf{L}ast},\textit{\textbf{O}utside},\textit{\textbf{U}nit}) instead of a state-agnostic token tagging approach.

\textbf{Slim model}: Without the use of a transfer-learning-based approach, in \textit{SpaCy} the transformation from discrete tokens into a dense vector representation is implemented by a model that is usually trained from scratch. Such model includes the embedding of the tokens into vectors via Bloom embeddings and further uses convolutional and dense layers to establish context-awareness and feature abstraction.

\textbf{Transfer-learning}: Inspired by the success of transformer-based neural networks and their effectiveness on language modeling through pre-training on large scale text corpora, transfer-learning-based methods using deep transformer models can also contribute to stronger entity recognition performance by providing contextualized token embeddings through earlier pre-training without the need to train such large models from scratch. As one instance, the masked language model BERT and several descendants have been released with pre-trained weights for various different languages including German, making it well-suited for transfer-learning.

\textbf{Entity Parsing}: The entity parsing is trained on top of the token vectors from a slim model or a pre-trained language model. First, an additional feature vector per token is predicted by a dense layer. During document parsing in NER, each token is processed depending on the parsing state. The state vector at a certain position is composed of the indices of the current token, the first token of the last entity and the previous token. The state-specific feature vector is compiled by adding up the token-specific feature vectors from the indices of the state vector. Given this feature vector, the next action can be estimated by a dense layer while invalid actions are masked by setting their probabilities to zero. The action determines the annotation of the current token and the next state when parsing the next token.

\section{Results}
\subsection{Dataset Acquisition}
The source dataset from the \textit{2018 n2c2 shared task on ADE and medication extraction in EHR} consists of 404 annotated text documents. The annotation includes the labels \textit{Strength}, \textit{Form}, \textit{Dosage}, \textit{Route}, \textit{Frequency}, \textit{Drug}, \textit{Duration}, \textit{Reason}, \textit{ADE}. The documents are split into sentences using the SpaCy sentencizer for English texts. After translation and word alignment estimation we obtain our raw dataset of 17938 German sentences. The annotation distribution of the raw dataset is shown in Table \ref{tab:rawdataset}.
\begin{table}[!ht]
    \centering
    \begin{tabular}{l|r}
        \hline
        \textbf{NER Tag} & \textbf{Count} \\
        \hline
        \hline
        Drug         & 26003       \\
        Route        & 8560        \\
        Reason       & 6244        \\
        Strength     & 10546       \\
        Frequency    & 9794        \\
        Duration     & 956         \\
        Form         & 10546       \\
        Dosage       & 6700        \\
        ADE          & 1557        \\
        \hline
    \end{tabular}
    \caption{The distribution of annotations in the (raw) synthesized dataset in absolute numbers. Note that a single tag sample count may include multiple tokens. The dataset consists of 16632 sentences.}
    \label{tab:rawdataset}
\end{table}

For further clean-up of the raw dataset, sentences that do not contain any entity label are discarded from the set of sentences, resulting in a total of 16632 sentence samples. Due to the nature of the labels \textit{ADE} and \textit{Reason} of being rather weakly defined and not of broad use for general applicability, we decided to drop these annotation labels from the dataset. As additional clean-up step, the annotation label \textit{Route} was removed as well since we concluded that the label samples lack of textual diversity after a swift investigation on the actual label occurrences.

\subsection{Entity Recognition Training}
Before the entity recognition model is trained, we split the previously described, filtered dataset into training, validation and test set (80\%,10\%,10\%). The split statistics are provided in Table \ref{tab:cleaneddataset}. Since the IOB-based entity recognition parser requires the annotated dataset to contain only non-overlapping annotation spans, annotation overlaps are resolved by removing the annotation span of shorter length while only preserving the longest span.
\begin{table}[!ht]
    \centering
    \begin{tabular}{l|r|r|r|r}
        \hline
        \textbf{Dataset} & \textbf{Split} & \textbf{\# Tokens} & \textbf{\# Entities} & \textbf{\# Sentences} \\
        \hline
        \hline
        Train Set      & 0.8 & 293693 & 50955 & 13306 \\
        Validation Set & 0.1 & 37218  & 6420  & 1663  \\
        Test Set       & 0.1 & 36168  & 6064  & 1663  \\
        \hline
        Total          & 1.0 & 367079 & 63439 & 16632 \\
        \hline
    \end{tabular}
    \caption{Information on the filtered dataset. Overlapping annotation spans were removed. The following NER tags were omitted: Route, Reason, ADE}
    \label{tab:cleaneddataset}
\end{table}

We investigate the ability of improving the entity recognition performance by the means of transfer-learning on deep language models on the basis of two German models:
\begin{itemize}
    \item \textbf{German BERT}\cite{chan_german_2019}\footnote{\url{https://www.deepset.ai/german-bert}} (\textit{bert-base-german-cased}): The model from Deepset AI follows the default architecture of BERT and has been specifically pre-trained on German data. The pre-training dataset stems from German Wikipedia, OpenLegalData, and German news articles.
    \item \textbf{GottBERT}\cite{schreible_gottbert_2020}: The model is based on the RoBERTa architecture and has been trained on the OSCAR dataset using the fairseq implementation. OSCAR is a German subset of CommonCrawl.
\end{itemize}

Both language models are publicly available. We retrieve both models from the Huggingface platform. For fine-tuning the entity recognizer on top of the language model, we utilize SpaCy for training. In this context, the model-specific tokenizer is inherited from the language model.

The training was performed on a single Nvidia Titan RTX. The training took 8-47 minutes (\textit{German BERT}: 47m, \textit{GottBERT}: 26m, \textit{Slim}: 8m).
In order to measure the differences in performance scores, we also compare the SpaCy Slim model using the same training and test set as baseline model, as well as the publicly available GERNERMED model as static model evaluated on the test set. It should be noted that the GERNERMED model scores must be considered as tainted because its weights are trained on a dataset that might partially contain samples from our test set. For evaluation, the NER procedure is considered as a token-wise multi-class classification problem. We computed the precision (\textit{Pr}), recall (\textit{Re}) and F1-score (\textit{F1}) for each individual label class as well as its respective (class-frequency-weighted) average score (\textit{Total}).
The final results on the test set are depicted in Table \ref{tab:eval1}.

\begin{table}[!t]
    \centering
    \begin{tabular}{|ll|rrrrrr|r|}
        \hline
        \multicolumn{2}{|l|}{\textit{\textbf{Scores on Test Set}}} & \multicolumn{6}{|c|}{\textbf{NER Tags}} & \textbf{} \\
        \hline
        \textbf{Model} & & \textbf{Str} & \textbf{Dur} & \textbf{Form} & \textbf{Dos} & \textbf{Drug} & \textbf{Freq} & \textbf{Total} \\
        \hline
        \hline
        \multirow{3}{*}{\shortstack[l]{GERNERMED++\\(GottBERT)}}
        & Pr & \textbf{0.971} & {0.806} & {0.947} & \textbf{0.967} & \textbf{0.969} & \textbf{0.880} & \textbf{0.942} \\
        & Re & {0.964} & \textbf{0.825} & \textbf{0.969} & \textbf{0.971} & {0.923} & {0.953} & \textbf{0.950} \\
        & F1 & \textbf{0.967} & \textbf{0.815} & {0.958} & {0.969} & {0.945} & \textbf{0.915} & \textbf{0.946} \\
        \hline
        \multirow{3}{*}{\shortstack[l]{GERNERMED++\\(GermanBERT)}}
        & Pr & {0.944} & {0.791} & {0.956} & {0.963} & \textbf{0.969} & {0.859} & {0.932} \\
        & Re & \textbf{0.973} & \textbf{0.825} & {0.962} & \textbf{0.971} & \textbf{0.933} & {0.924} & {0.947} \\
        & F1 & {0.958} & {0.807} & \textbf{0.959} & {0.967} & \textbf{0.951} & {0.890} & {0.939} \\
        \hline
        \multirow{3}{*}{\shortstack[l]{GERNERMED++\\(SpaCy Slim)}}
        & Pr & {0.965} & \textbf{0.823} & \textbf{0.965} & {0.958} & {0.929} & {0.855} & {0.926} \\
        & Re & {0.967} & {0.749} & {0.950} & \textbf{0.971} & {0.884} & \textbf{0.966} & {0.941} \\
        & F1 & {0.966} & {0.784} & {0.957} & {0.964} & {0.906} & {0.907} & {0.932} \\
        \hline
        \multirow{3}{*}{GERNERMED$^{\dagger}$}
        & Pr & {0.916} & {0.613} & {0.842} & {0.915} & {0.644} & {0.739} & {0.790} \\
        & Re & {0.917} & {0.697} & {0.882} & {0.959} & {0.634} & {0.901} & {0.841} \\
        & F1 & {0.917} & {0.652} & {0.861} & {0.937} & {0.639} & {0.812} & {0.814} \\
        \hline
        \multicolumn{9}{|l|}{\small Note: $^{\dagger}$ specific training set might be tainted by samples from the test set.} \\
        \hline
    \end{tabular}
    \caption{Evaluation of models' performance scores on test set.}
    \label{tab:eval1}
\end{table}

Both transfer-learning-based approaches exhibit strong performance in absolute numbers. Though to our surprise, German BERT achieves significantly inferior performance scores in direct comparison to GottBERT. We attribute this performance gap to the differences in pre-training dataset sizes for German BERT (12GB) and GottBERT (145GB).

\subsection{Out-of-Distribution Evaluation}
The evaluation on the test set does not provide valuable information on how a model can maintain its scores beyond the scope of the train and test set. A known property of neural networks as statistical models is their ability to overfit to the training dataset. While strong performance on the test set indicates the ability to abstract from individual samples without blunt sample memorization, it cannot measure the model's reliance on the inherent bias of the dataset and its ability to generalize to \textit{out-of-distribution}(OoD) samples. To investigate the OoD generalization ability, we retrieved 30 text samples provided from independent physicians annotated with equivalent labels to our dataset and evaluated the models' performance on this separated dataset. Since the physicians were instructed to use the class labels from our initial dataset, the OoD samples are annotated with matching label classes and can be directly used for full evaluation of our models. The results are shown in Table \ref{tab:evalOoD}.

\begin{table}[!ht]
    \centering
    \begin{tabular}{|ll|rrrrrr|r|}
        \hline
        \multicolumn{2}{|l|}{\textit{\textbf{Scores on OoD Dataset}}} & \multicolumn{6}{|c|}{\textbf{NER Tags}} & \textbf{} \\
        \hline
        \textbf{Model} & & \textbf{Str} & \textbf{Dur} & \textbf{Form} & \textbf{Dos} & \textbf{Drug} & \textbf{Freq} & \textbf{Total} \\
        \hline
        \hline
        \multirow{3}{*}{\shortstack[l]{GERNERMED++\\(GottBERT)}}
        & Pr & {0.866} & \textbf{1.000} & \textbf{1.000} & \textbf{0.125} & \textbf{0.891} & \textbf{0.923} & \textbf{0.883} \\
        & Re & \textbf{0.960} & {0.400} & \textbf{0.632} & \textbf{0.250} & {0.932} & {0.615} & \textbf{0.835} \\
        & F1 & \textbf{0.911} & {0.571} & \textbf{0.774} & \textbf{0.167} & \textbf{0.911} & \textbf{0.738} & \textbf{0.845} \\
        \hline
        \multirow{3}{*}{\shortstack[l]{GERNERMED++\\(GermanBERT)}}
        & Pr & \textbf{0.955} & \textbf{1.000} & {0.909} & {0.077} & {0.830} & {0.456} & {0.817} \\
        & Re & {0.832} & \textbf{0.800} & {0.526} & \textbf{0.250} & \textbf{1.000} & \textbf{0.667} & {0.797} \\
        & F1 & {0.889} & \textbf{0.889} & {0.667} & {0.118} & {0.907} & {0.542} & {0.794} \\
        \hline
        \multirow{3}{*}{\shortstack[l]{GERNERMED++\\(SpaCy Slim)}}
        & Pr & {0.951} & {0.000} & \textbf{1.000} & {0.111} & {0.690} & {0.486} & {0.778} \\
        & Re & {0.772} & {0.000} & {0.316} & \textbf{0.250} & {0.659} & {0.462} & {0.623} \\
        & F1 & {0.852} & {0.000} & {0.480} & {0.154} & {0.674} & {0.474} & {0.679} \\
        \hline
        \multirow{3}{*}{GERNERMED}
        & Pr & {0.851} & {0.000} & {0.500} & {0.045} & {0.460} & {0.390} & {0.619} \\
        & Re & {0.624} & {0.000} & {0.158} & {0.250} & {0.523} & {0.410} & {0.500} \\
        & F1 & {0.720} & {0.000} & {0.240} & {0.077} & {0.489} & {0.400} & {0.541} \\
        \hline\hline
        \multicolumn{2}{|l|}{\textbf{\#Labels}} & 37 & 3 & 19 & 4 & 36 & 20 & {119}\\
        \hline
    \end{tabular}
    \caption{Evaluation of models' performance scores on separated OoD dataset.}
    \label{tab:evalOoD}
\end{table}

The results display the impact of the transfer-learning-based NER models in order to preserve strong performance on OoD data samples. However similar to the results on the test set, German BERT performs inferior to the GottBERT-based model by an increased margin according to the weighted F1-score. In contrast, the baseline models suffer from substantially degraded scores in comparison to their scores on the test set.

Due to the sparseness and independent origin of the OoD dataset, the number of labels is imbalanced across individual class labels and explains that the evaluation scores can yield \textit{1.0} or \textit{0.0} in several situations.

\subsection{Related Datasets}

We select three relevant datasets in order to further evaluate our models. To put our results in perspective, we also evaluate the reference model from GGPONC\cite{borchert_ggponc_2022} on these datasets. The entity labels from the datasets differ from the labels of our training dataset and our OoD dataset and limits our ability to perform a complete comparison of our model with respect to all label classes. All related datasets provide annotation information on entities which we consider to be semantically strongly related to the class label \textit{Drug}, although the datasets commonly lack clear definitions on their label classes. We evaluate the scores as a classification task on token- and character-level. The results are shown in Table \ref{tab:evalRelated}.

\begin{table}[!ht]
    \centering
    \resizebox{0.7\textwidth}{!}{%
    \begin{tabular}{|ll|rr|}
        \hline
        \multicolumn{2}{|l|}{\textit{\textbf{Scores on Related Datasets}}} & \multicolumn{2}{|c|}{\textbf{F1 Scores}} \\
        \hline
        \textbf{Model / Dataset} & & \textbf{Drug (char-wise)} & \textbf{Drug (token-wise)} \\
        \hline
        \hline
        \multicolumn{2}{|c|}{\textbf{Medline Dataset\cite{kors_multilingual_2015}}} & \multicolumn{2}{|c|}{\textbf{Drug=CHEM}} \\
        \hline
        \multirow{3}{*}{\shortstack[l]{GERNERMED++\\(GottBERT)}}
        & Pr & {0.858} & {0.837} \\
        & Re & \textbf{0.701} & \textbf{0.706} \\
        & F1 & \textbf{0.772} & {0.766} \\
        \hline
        \multirow{3}{*}{\shortstack[l]{GERNERMED++\\(GermanBERT)}}
        & Pr & \textbf{0.885} & \textbf{0.875} \\
        & Re & {0.638} & {0.686} \\
        & F1 & {0.742} & \textbf{0.769} \\
        \hline
        \multirow{3}{*}{\shortstack[l]{GERNERMED++\\(SpaCy Slim)}}
        & Pr & {0.437} & {0.500} \\
        & Re & {0.182} & {0.216} \\
        & F1 & {0.257} & {0.301} \\
        \hline
        \multirow{3}{*}{\shortstack[l]{GERNERMED}}
        & Pr & {0.477} & {0.414} \\
        & Re & {0.207} & {0.235} \\
        & F1 & {0.288} & {0.300} \\
        \hline
        \multirow{3}{*}{\shortstack[l]{GGPONC\cite{borchert_ggponc_2022}}}
        & Pr & {0.822} & {0.771} \\
        & Re & {0.488} & {0.529} \\
        & F1 & {0.612} & {0.628} \\
        \hline
        \multicolumn{2}{|c|}{\textbf{GGPONC Dataset\cite{borchert_ggponc_2022}}} & \multicolumn{2}{|c|}{\textbf{Drug=Chemicals\_Drugs}} \\
        \hline
        \multirow{3}{*}{\shortstack[l]{GERNERMED++\\(GottBERT)}}
        & Pr & {0.535} & {n/a} \\
        & Re & {0.664} & {n/a} \\
        & F1 & {0.592} & {n/a} \\
        \hline
        \multirow{3}{*}{\shortstack[l]{GERNERMED++\\(GermanBERT)}}
        & Pr & {0.522} & {n/a} \\
        & Re & {0.645} & {n/a} \\
        & F1 & {0.577} & {n/a} \\
        \hline
        \multirow{3}{*}{\shortstack[l]{GERNERMED++\\(SpaCy Slim)}}
        & Pr & {0.185} & {n/a} \\
        & Re & {0.433} & {n/a} \\
        & F1 & {0.260} & {n/a} \\
        \hline
        \multirow{3}{*}{\shortstack[l]{GERNERMED}}
        & Pr & {0.089} & {n/a} \\
        & Re & {0.303} & {n/a} \\
        & F1 & {0.138} & {n/a} \\
        \hline
        \multirow{3}{*}{\shortstack[l]{GGPONC\cite{borchert_ggponc_2022}}}
        & Pr & \textbf{0.636} & {n/a} \\
        & Re & \textbf{0.737} & {n/a} \\
        & F1 & \textbf{0.683} & {n/a} \\
        \hline
        \multicolumn{2}{|c|}{\textbf{BRONCO Dataset\cite{kittner_annotation_2021}}} & \multicolumn{2}{|c|}{\textbf{Drug=MEDICATION}} \\
        \hline
        \multirow{3}{*}{\shortstack[l]{GERNERMED++\\(GottBERT)}}
        & Pr & {0.673} & {0.726} \\
        & Re & \textbf{0.789} & \textbf{0.752} \\
        & F1 & \textbf{0.726} & \textbf{0.739} \\
        \hline
        \multirow{3}{*}{\shortstack[l]{GERNERMED++\\(GermanBERT)}}
        & Pr & \textbf{0.684} & \textbf{0.730} \\
        & Re & {0.677} & {0.637} \\
        & F1 & {0.680} & {0.680} \\
        \hline
        \multirow{3}{*}{\shortstack[l]{GERNERMED++\\(SpaCy Slim)}}
        & Pr & {0.320} & {0.378} \\
        & Re & {0.512} & {0.486} \\
        & F1 & {0.394} & {0.425} \\
        \hline
        \multirow{3}{*}{\shortstack[l]{GERNERMED}}
        & Pr & {0.155} & {0.148} \\
        & Re & {0.478} & {0.482} \\
        & F1 & {0.234} & {0.227} \\
        \hline
        \multirow{3}{*}{\shortstack[l]{GGPONC\cite{borchert_ggponc_2022}}}
        & Pr & {0.573} & {0.346} \\
        & Re & {0.449} & {0.430} \\
        & F1 & {0.504} & {0.384} \\
        \hline
    \end{tabular}
    }
    \caption{Evaluation of models' F1 scores on related dataset. The GGPONC reference model is evaluated for comparison. To allow fair comparison, only Drug-related label classes are selected. Annotations from the GGPONC dataset do not align onto the tokens from the SpaCy tokenizer and are therefore omitted.}
    \label{tab:evalRelated}
\end{table}

To no suprise, the GGPONC reference model archives better performance on its native GGPONC dataset, yet all our models with transfer-learning-based, pre-trained BERT encoder outperform the reference model, our slim model and the baseline GERNERMED model significantly. Considering that the baseline GGPONC model was developed in traditional fashion using a manually crafted German dataset, the archived performance margins from both GottBERT- and GermanBERT-based models are unexpected. Throughout the tasks, the GottBERT-based model beats the GermanBERT-based model which is consistent with previous observations.

\section{Discussion}
Our results indicate strong performance of all models on the test set, however our evaluation on the OoD dataset as well as on external, related datasets shows the impact of using the transfer-learning abilities of pre-trained BERT-based feature encoders to solidify the robust performance on such external datasets. Considering the fact that our models were developed without additional manual work of annotating datasets and only a public non-German dataset was used, the obtained models compete surprisingly well with the pre-existing reference model and are able to outperform it on independent datasets. The lack of more independent annotated datasets, lacking matching annotation labels and unclear label class definitions still limit the possibility to deeper evaluate and compare novel models and methods.

In general, considering the current poor availability of open medical NLP systems for non-English natural language as well as for German in particular, our refined approach demonstrates a powerful opportunity to build a strong medical NER model solely by the use of a public English dataset. Another notable factor of the use of public datasets is its inherent immunity against privacy threats from adversarial training data extraction, which allows us to publish our resulting models for third party usage. In contrast, NLP systems trained on internal data from proprietary sources like local clinics are often kept unavailable due to legal reasons.

\section{Conclusion}
In this work, we presented a fine-tuned German NER model for medical entities using deep pre-trained language models by the means of transfer-learning and demonstrated its ability to outperform the basic baseline model on the test set and on an out-of-distribution dataset. In comparison to the existing GGPONC reference model, we showed competitive results on external datasets and outperformed the reference model on all independent datasets. Furthermore, we described the process and its relevant improvements to obtain a medical-specific German dataset. No proprietary, internal data was used for training. Our open NER model is publicly available for third-party use.

\section*{Acknowledgment}
This work is a part of the DIFUTURE project funded by the German Ministry of Education and Research (Bundesministerium für Bildung und Forschung, BMBF) grant FKZ01ZZ1804E.

\bibliographystyle{plain}

\end{document}